\documentclass[sn-mathphys,Numbered]{sn-jnl}

\usepackage{silence}
\usepackage{graphicx}%
\usepackage{subcaption}
\usepackage{multirow}%
\usepackage{amsmath,amssymb,amsfonts}%
\usepackage{amsthm}%
\usepackage{mathrsfs}%
\usepackage[title]{appendix}%
\usepackage[dvipsnames]{xcolor}%
\usepackage{textcomp}%
\usepackage{manyfoot}%
\usepackage{booktabs}%
\usepackage{algorithm}%
\usepackage{algorithmicx}%
\usepackage{algpseudocode}%
\usepackage{listings}%
\usepackage{array}%
\usepackage{adjustbox}%
\usepackage[edges]{forest}%
\usepackage[mode=buildnew]{standalone}%
\usepackage{microtype}
\usepackage[inline,shortlabels]{enumitem}%
\usepackage{xspace}
\usepackage{makecell}
\usepackage{anyfontsize}
\usepackage{tabularx}
\usepackage{bookmark}
\usepackage{dirtree}
\usepackage{pifont}
\usepackage{orcidlink}
\usepackage{tcolorbox}
\usepackage{hyperref}

\newcommand{\cmark}{\ding{51}}%
\newcommand{\xmark}{\ding{55}}%

\newcommand{\pedestrian}{PEDESTRIAN\xspace}
\newcommand{\taxLevel}[1]{\textit{#1}}
\newcommand{\model}[1]{\texttt{#1}\xspace}

\begin{document}

\title[Article Title]{PEDESTRIAN: An Egocentric Vision Dataset for Obstacle Detection on Pavements}


\author*[1,2]{\fnm{Marios} \sur{Thoma} \orcidlink{0000-0001-7364-5799}}\email{m.thoma@cyens.org.cy}
\equalcont{These authors contributed equally to this work.}

\author[1,3]{\fnm{Zenonas} \sur{Theodosiou} \orcidlink{0000-0003-3168-2350}}\email{zenonas.theodosiou@cut.ac.cy}
\equalcont{These authors contributed equally to this work.}

\author[4]{\fnm{Harris} \sur{Partaourides} \orcidlink{0000-0002-8555-260X}}\email{harris.partaourides@ethicalaicy.com}

\author[1]{\fnm{Vassilis} \sur{Vassiliades} \orcidlink{0000-0002-1336-5629}}\email{v.vassiliades@cyens.org.cy}

\author[2,1]{\fnm{Loizos} \sur{Michael }}\email{loizos@ouc.ac.cy}

\author[1,5]{\fnm{Andreas} \sur{Lanitis} \orcidlink{0000-0001-6841-8065}}\email{andreas.lanitis@cut.ac.cy}

\affil*[1]{\orgname{CYENS Centre of Excellence}, \orgaddress{\street{Dimarchou Lellou Demetriadi 23}, \city{Nicosia}, \postcode{1016}, \country{Cyprus}}}

\affil[2]{\orgname{Open University Cyprus}, \orgaddress{\street{Giannou Kranidioti 33}, \city{Nicosia}, \postcode{2220}, \country{Cyprus}}}

\affil[3]{\orgdiv{Department of Communication and Internet Studies}, \orgname{Cyprus University of Technology}, \orgaddress{\street{Archiepiskopou Kyprianou 30}, \city{Limassol}, \postcode{3036}, \country{Cyprus}}}

\affil[4]{\orgname{AI Cyprus Ethical Novelties Ltd}, \orgaddress{\street{Andrea Zakou 4}, \city{Limassol}, \postcode{3095}, \country{Cyprus}}}

\affil[5]{\orgdiv{Department of Multimedia and Graphic Arts}, \orgname{Cyprus University of Technology}, \orgaddress{\street{Archiepiskopou Kyprianou 30}, \city{Limassol}, \postcode{3036}, \country{Cyprus}}}

\abstract{%
    Walking has always been a primary mode of transportation and is recognized as an essential activity for maintaining good health.
    Despite the need for safe walking conditions in urban environments, sidewalks are frequently obstructed by various obstacles that hinder free pedestrian movement.
    Any object obstructing a pedestrian's path can pose a safety hazard.
    The advancement of pervasive computing and egocentric vision techniques offers the potential to design systems that can automatically detect such obstacles in real time, thereby enhancing pedestrian safety.
    The development of effective and efficient identification algorithms relies on the availability of comprehensive and well-balanced datasets of egocentric data.
    In this work, we introduce the \pedestrian dataset, comprising egocentric data for 29 different obstacles commonly found on urban sidewalks.
    A total of 340 videos were collected using mobile phone cameras, capturing a pedestrian's point of view.
    Additionally, we present the results of a series of experiments that involved training several state-of-the-art deep learning algorithms using the proposed dataset, which can be used as a benchmark for obstacle detection and recognition tasks.
    The dataset can be used for training pavement obstacle detectors to enhance the safety of pedestrians in urban areas.
}


\maketitle

\section{Introduction}

Walking is a popular form of physical exercise, known for its positive impact on health.
Being an aerobic and bone-strengthening activity, it can reduce the risk of chronic diseases and promote an active lifestyle~\cite{piercy2018ActivityGuidelines}.
Despite the widespread promotion of the benefits of walking by governments and health organizations, pedestrian safety continues to be a major concern, as road accidents pose a significant safety risk.
The well-being of road users is influenced by numerous factors, such as speeding and rule violations by drivers, pedestrian negligence, and poor road conditions.
Such conditions can be particularly hazardous for vulnerable groups like the elderly and disabled~\cite{haghighi2020ChallengesIranian}.

Poor road conditions, including inadequate lighting, poorly-maintained road surfaces, and obstructed pedestrian sidewalks are recognized to be some of the major factors contributing to road accidents~\cite{angin2021analysis}.
Several studies have analyzed the behavior of pedestrians when faced with obstacles in their path.
For example, in a study conducted by \citet{ding2020Evacuation}, the impact of different obstacle placements on pedestrian evacuation was examined.
The results showed that the presence of obstacles in pedestrian pathways resulted in a significant delay before pedestrians could maneuver past the obstacles.
Recent advancements in computer vision and mobile computing have paved the way for various applications aimed at enhancing the safety and well-being of pedestrians.
Egocentric vision is a novel approach that utilizes data collected through wearable cameras or smartphones to enable real-time recognition of obstacles hindering safe walking within urban areas.

The work presented in this paper is a continuation of our previous work in this area~\cite{thoma2023PerformanceAssessment}, where automatic recognition of obstacles present in pedestrian pathways was addressed, focusing on the analysis of the performance of deep learning models under different conditions.
However, to develop effective systems and applications for smartphones, large egocentric datasets are needed for developing the machine learning models~\cite{Theodosiou2022Artworks}.
Compared to our previous work in the area, the egocentric dataset was extended and, in this work, we present the \pedestrian dataset, consisting of 340 videos collected on urban sidewalks and focusing on several types of obstacles impeding the safe movement of pedestrians.
In total the \pedestrian dataset includes 29 obstacle types, whereas in our previous work the dataset included 9 types.
The data was collected by a smartphone user while walking around the city of Nicosia, Cyprus.
Using the mobile camera, videos of the obstacles were collected from a pedestrian's point of view.
In addition to introducing the dataset, the performance of several state-of-the-art deep learning neural networks in obstacle detection using the dataset is also evaluated, setting benchmark standards for detection performance.
The \pedestrian dataset and related resources have been made publicly available to encourage further research in the area of obstacle detection on pavements, and the development of related applications.

The remainder of the paper proceeds with Section~\ref{sec:litSurvey} providing an overview of the state-of-the-art research in the field of obstacle recognition and available egocentric datasets.
Section~\ref{sec:methodology} details the entire process of the dataset creation.
In Section~\ref{sec:benchmarks}, the experimental procedure for evaluating the performances of benchmark tests, and the insights obtained from the evaluation are discussed.
Finally, Section~\ref{sec:conclusion} summarizes our work, highlights its contributions, and suggests avenues for future enhancements.

\section{Related Work} \label{sec:litSurvey}

Egocentric vision is a perceptual method that involves the use of wearable cameras attached to a person to capture egocentric images or videos~\cite{mann2014Introduction}.
The advent of automatic egocentric data analysis has led to a plethora of applications designed to enhance daily human activities, not the least among them the augmentation of pedestrian safety.
The use of Deep Learning (DL) is now the dominant approach for analyzing egocentric data.
Thus, a wide range of egocentric datasets has been created and made publicly available for the training of DL models.

\subsection{Pedestrian Safety}

Various technologies have been developed with the aim to improve pedestrian safety.
One such technology is ObstacleWatch, an obstacle detection model designed by \citet{wang2018ObstacleWatch}, which uses acoustic signals emitted by smartphone speakers to determine the distance between the user and an obstacle.
However, this approach proved to be inefficient due to the noise present in public spaces.
Another application called LookUp was proposed by \citet{jain2015LookUp}, which uses shoe-mounted inertial sensors to detect transitions from pedestrian walkways onto the road to alert texting pedestrians, achieved a detection rate of 90\%.
\citet{liu2017InfraSee} developed a solution named InfraSee, which utilizes infrared sensors mounted on smartphones to detect hazardous situations and alerts the user.
These approaches, however, necessitate additional equipment such as shoe-mounted sensors or infrared sensors, rendering them less practical for widespread adoption among the general public due to the extra hardware requirements and the inconvenience of carrying or wearing additional devices.

In contrast, smartphone-based applications offer a more practical and accessible solution for enhancing pedestrian safety without the need for supplementary equipment.
WalkSafe, developed by \citet{wang2012Walksafe}, is a smartphone-based application that uses the smartphone's back camera to detect oncoming vehicles and alerts the user during active phone calls.
The application uses decision trees to classify images and achieved an efficiency rate of 77\%.
A similar application, called Inspector, developed by \citet{tang2016DistractedMobileUsers}, alerts users when they are approaching the edge of pedestrian areas.
The model uses simple keypoint detection and K-means clustering for feature extraction and uses Normal Bayes and K-nearest neighbors models to classify images with an accuracy rate of 92\%--99\%.
TerraFirma, an application proposed by \citet{jain2019RecognizingTextures}, takes a different approach, choosing to identify the material composition of pedestrian walkways, instead of obstacles on the walkways themselves, in order to warn pedestrians when they transition from walkways to the street.
Support Vector Machine classifiers were used to identify various ground surface types with an accuracy rate of 90\%.
Another application, AutoADAS, proposed by \citet{wei2015accidentDetection}, uses smartphone cameras to detect objects in the environment and measure their distance from the user, warning them in case a potential collision is predicted.
The application utilizes the user's behavior profile collected from sensors on mobile devices, making it more personalized to the user.
Similarly, \citet{foerster2014SpareEye} developed SpareEye, an Android app that detects changes in the background of a mobile camera's video stream and notifies the user when the distance between objects is reduced in each frame.

\citet{hasan2022PedestrianIoT} presented a comprehensive review of existing pedestrian safety models, highlighting the limitations and potential areas for further improvement.
One of the issues highlighted was that, although various egocentric datasets are available, relatively few datasets focusing on pedestrian obstacle detection exist.
One of the few such datasets available is that created by \citet{theodosiou2020FirstPersonDatabase}, comprised of images of obstacles captured from a pedestrian's perspective.
The dataset was utilized to fine-tune a \model{VGG16} model~\cite{simonyan2015VeryDeep}, achieving a training accuracy of 65\% and validation accuracy of 55\%.
Subsequently, the fine-tuned \model{VGG16} model was embedded in a smartphone application that aided pedestrians in capturing images of barriers they encountered and report their geographic location to the authorities, in order to enhance pedestrian safety~\cite{thoma2021Smartphone}.
Additionally, in \citet{theodosiou2022BarrierDetection}, three combinations of object detection models were trained using the aforementioned pedestrian-based image dataset, namely
a \model{Faster Region-Based CNN}~\cite{ren2017FasterRCNN} using \model{Inception-v2}~\cite{szegedy2016RethinkingInception} and \model{ResNet-50}~\cite{he2015DeepResidual} for feature extraction, and a \model{Single Shot MultiBox Detector (SSD)}~\cite{liu2016SSD} using \model{MobileNetV2}~\cite{sandler2018MobileNetV2} for feature extraction.
The models achieved, respectively, an average accuracy of 88.4\%, 81.7\% and 75.6\%, demonstrating the dataset's efficacy in training models to detect pedestrian obstacles on sidewalks.

\subsection{Egocentric Visual Datasets}

Several egocentric vision datasets have been developed for a variety of tests and applications, as depicted in Table~\ref{tab:Available_datasets}.
EPIC-Kitchens~\cite{damen2022EpicKitchens} is a benchmark egocentric video dataset focusing on kitchen activities.
It consists of 100 hours of video data captured by head mounted cameras, while participants were performing various cooking tasks in kitchen environments.
The dataset has been used for different purposes, such as action recognition, action detection and action anticipation.
Ego4D~\cite{grauman2022ego4d} is a large-scale egocentric video dataset containing over 3670 hours of daily-life video captured from a wearable camera in diverse indoor and outdoor environments.
The authors, in addition to the dataset, also proposed a series of benchmark tests for recall of past, present, and future activities.
Charades-Ego~\cite{sigurdsson2018ActorObserver} contains over 7500 files combining both first-person (actor) and third-person (observer) videos of daily activities, aiming to enable learning the connections between the two perspectives.
EGTEA Gaze+~\cite{li2018EyeBeholder} consists of 28 hours of video data collected using a head-worn camera focusing on both activities and gazes of the subjects.
The UT egocentric dataset~\cite{lee2012DiscoveringImportant} consists of 37 video hours collected through wearable cameras.
The collection concerns the daily lives of 4 different wearers and the dataset was initially used for video summarization based on the people and objects discovered in each event.

Different object categories are included in the EgoObjects dataset~\cite{zhu2023EgoObjects} which was created for object understanding in egocentric visual data.
The dataset covers 250 object categories in more than 9000 video files.
A large amount of video gestures were collected in the EgoGesture dataset~\cite{zhang2018EgoGesture}.
The gestures were gathered from a first-person point of view using 50 human subjects.
The TerraFirma dataset~\cite{jain2019RecognizingTextures} is concerned with images of sidewalks' surfaces from a pedestrian point of view.
The data was collected using smartphone cameras from volunteers walking in various cities.
In addition, other purpose datasets have been proposed, such as the DR(eye)Ve~\cite{palazzi2018Predicting} for driving, EGO-CH~\cite{ragusa2020EGOCH} for cultural sites, TREK-150~\cite{dunnhofer2023VisualObjTracking} for object tracking, etc.

Apart from the dataset presented in~\citet{theodosiou2020FirstPersonDatabase}, which had several limitations concerning its size, to our knowledge no other dataset has been proposed for the protection of pedestrians from the various obstacles that crowd sidewalks.
Although there are egocentric datasets that can be used to protect pedestrians, including datasets of objects, people driving and sidewalk surface textures, none focuses exclusively on the obstacles that obstruct the free movement and endanger the lives of pedestrians.

\begin{table*}[t]
    \centering
    \caption{Overview of available egocentric datasets.}
    \label{tab:Available_datasets}
    \resizebox{\textwidth}{!}{
        \begin{tabular}{lcclllc}
            \toprule
                                                             & \multicolumn{2}{c}{Camera} &            &                               &                               &                                                                                \\ \cmidrule(r){2-3}
            \multirow{-2}{*}[0.5ex]{Dataset}                 & Wearable                   & Smartphone & \multirow{-2}{*}[0.5ex]{Year} & \multirow{-2}{*}[0.5ex]{Size} & \multirow{-2}{*}[0.5ex]{Task}      & \multirow{-2}{*}[0.6ex]{\thead{Pedestrian \\Safety}}\\
            \midrule
            UT Egocentric~\cite{lee2012DiscoveringImportant} & \cmark                     & \xmark     & 2012                          & 37 video hrs                  & Video Summarization                & \xmark                                    \\
            Charades-Ego~\cite{sigurdsson2018ActorObserver}  & \cmark                     & \xmark     & 2018                          & 7860 videos                   & Object \& Action Recognition       & \xmark                                    \\
            EgoGesture~\cite{zhang2018EgoGesture}            & \cmark                     & \xmark     & 2018                          & 2000 videos                   & Hand Gesture Recognition           & \xmark                                    \\
            EGTEA Gaze+~\cite{li2018EyeBeholder}             & \cmark                     & \xmark     & 2018                          & 28 video hrs                  & Action \& Gaze Recognition         & \xmark                                    \\
            DR(eye)Ve~\cite{palazzi2018Predicting}           & \cmark                     & \xmark     & 2019                          & 6.17 video hrs                & Drivers' Focus Attention           & \cmark                                    \\
            TerraFirma~\cite{jain2019RecognizingTextures}    & \cmark                     & \cmark     & 2019                          & 9.25 video hrs                & Surface Texture Recognition        & \cmark                                    \\
            EGO-CH~\cite{ragusa2020EGOCH}                    & \cmark                     & \xmark     & 2020                          & 27 video hrs                  & Visitors' Behavioral Understanding & \xmark                                    \\
            \citet{theodosiou2020FirstPersonDatabase}        & \xmark                     & \cmark     & 2020                          & 1796 frames                   & Obstacle Detection                 & \cmark                                    \\
            TREK-150~\cite{dunnhofer2023VisualObjTracking}   & \cmark                     & \xmark     & 2021                          & 150 videos                    & Object Tracking                    & \xmark                                    \\
            Ego4D~\cite{grauman2022ego4d}                    & \cmark                     & \xmark     & 2022                          & 3670 video hrs                & Memory Recall                      & \xmark                                    \\
            EPIC-Kitchens~\cite{damen2022EpicKitchens}       & \cmark                     & \xmark     & 2022                          & 100 video hrs                 & Object \& Action Recognition       & \xmark                                    \\
            EgoObjects~\cite{zhu2023EgoObjects}              & \cmark                     & \xmark     & 2023                          & $>$9000 videos                & Object Understanding               & \xmark                                    \\
            \pedestrian                                      & \xmark                     & \cmark     & 2024                          & 82120 frames                  & Obstacle Detection                 & \cmark                                    \\
            \bottomrule
        \end{tabular}
    }
\end{table*}

\subsection{Obstacle Detection in Egocentric Datasets}

Various DL architectures, including Convolutional Neural Networks (CNNs), have been deployed with notable effectiveness in egocentric and mobile computing applications.
The constraints imposed by smartphone hardware necessitated the creation of the \model{MobileNetV2} framework~\cite{sandler2018MobileNetV2}, which is tailored specifically for these environments through the prioritization of compact model dimensions and reduced inference latency.
\model{MobileNetV2}, distinguished by its innovative inverted residual structure as detailed by \citet{sandler2018MobileNetV2}, represents a prominent advancement in CNN designs optimized for mobile platforms, offering enhanced efficiency in tasks such as image classification.
This architecture innovatively employs an expansion of low-dimensional features, followed by a reduction in dimensionality via depth-wise convolution, optimizing both computational resource usage and performance.

Despite the success of CNN-based models such as \model{MobileNetV2} (and its successors, \model{MobileNetV3-Small} and \model{MobileNetV3-Large}~\cite{howard2019MobileNetV3}), the pursuit for models that offer superior accuracy performance has led to the development of \model{Vision Transformer (ViT)} based architectures, as introduced by \citet{dosovitskiy2021ImageWorth}.
These architectures diverge from traditional CNN approaches by leveraging self-attention mechanisms~\cite{vaswani2017AttentionAllYouNeed}, a methodology that, rather than focusing on computational efficiency, aims to enhance the model's accuracy in various computer vision tasks, including image classification and object recognition.
The \model{ViT} architecture represents a significant shift in the field, positioning itself as a competitive alternative to the previously dominant CNN models, by potentially offering improvements in accuracy for certain applications.

\section{The \pedestrian Dataset} \label{sec:methodology}

\begin{figure}[t]
    \centering
    \includegraphics[width=\textwidth]{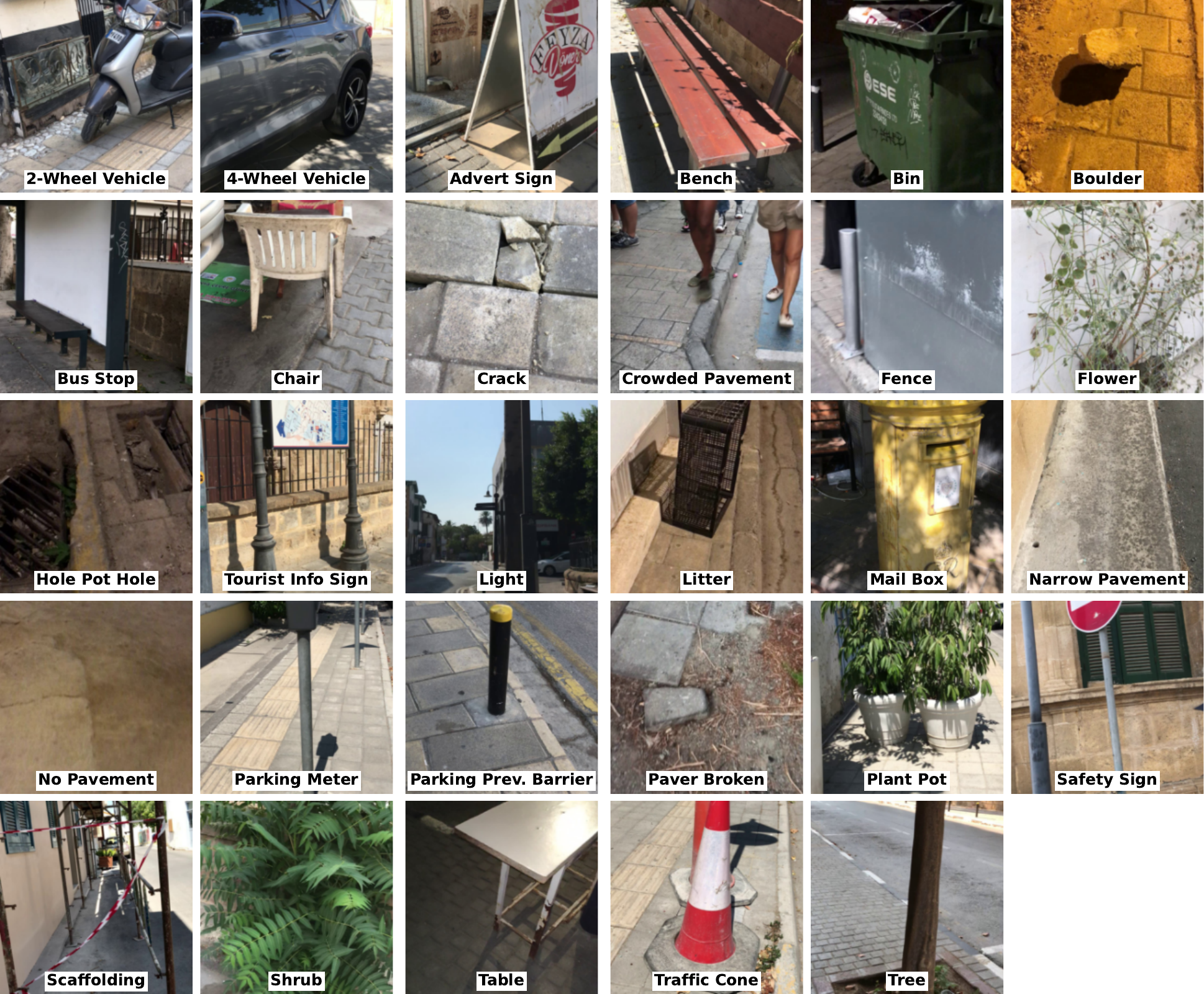}
    \caption{Example images for the 29 obstacle types in the \pedestrian dataset.}
    \label{fig:dataset-image-examples}
\end{figure}

\begin{figure}[t]
    \centering
    \includegraphics[width=1\textwidth]{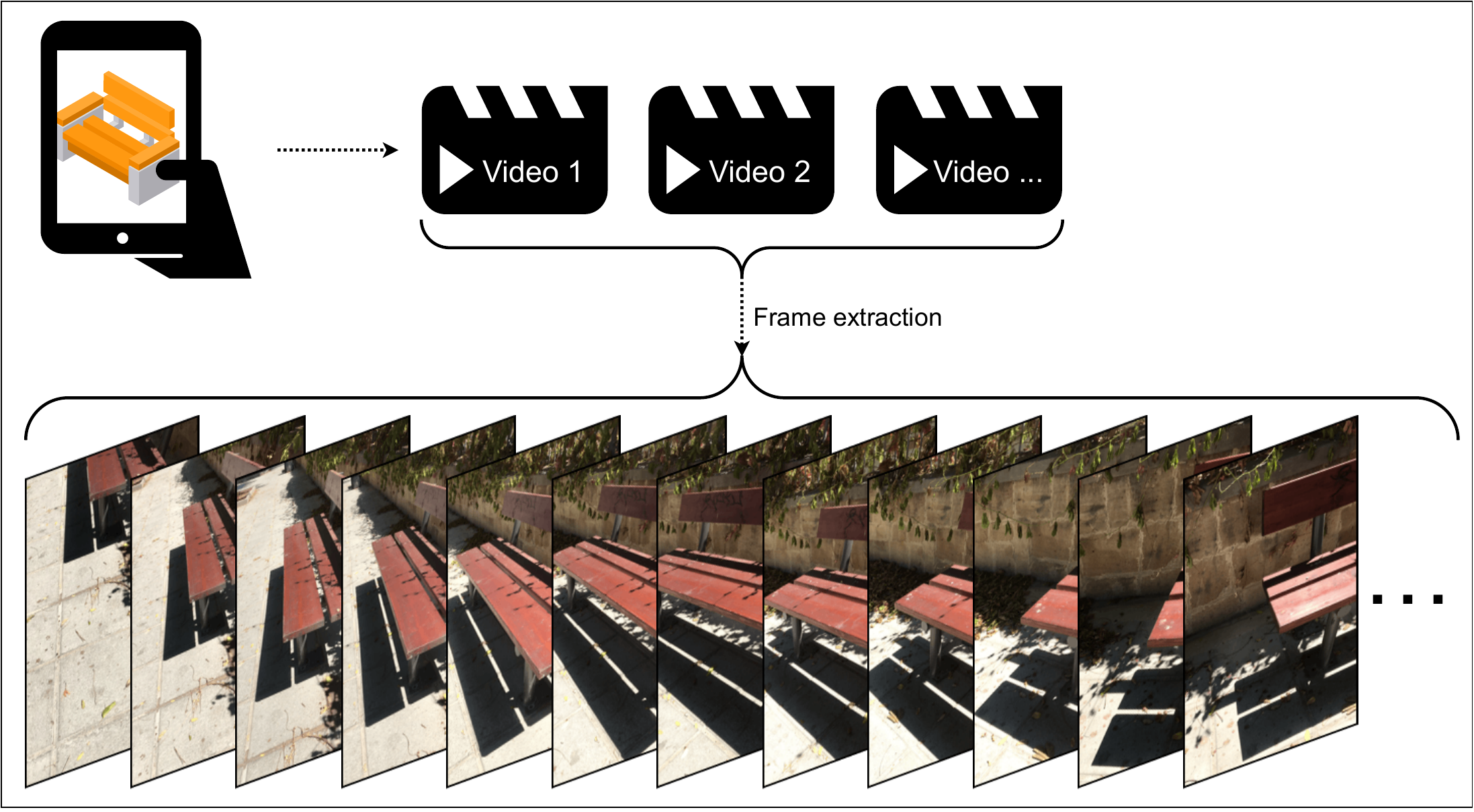}
    \caption{Pipeline for the generation of the \pedestrian dataset: multiple videos of each obstacle type were taken using a smartphone camera from a pedestrian's point of view, and then individual frames were extracted from the collected videos.}
    \label{fig:frame-extraction}
\end{figure}

\begin{figure}[t]
    \centering
    \includegraphics[width=1\textwidth]{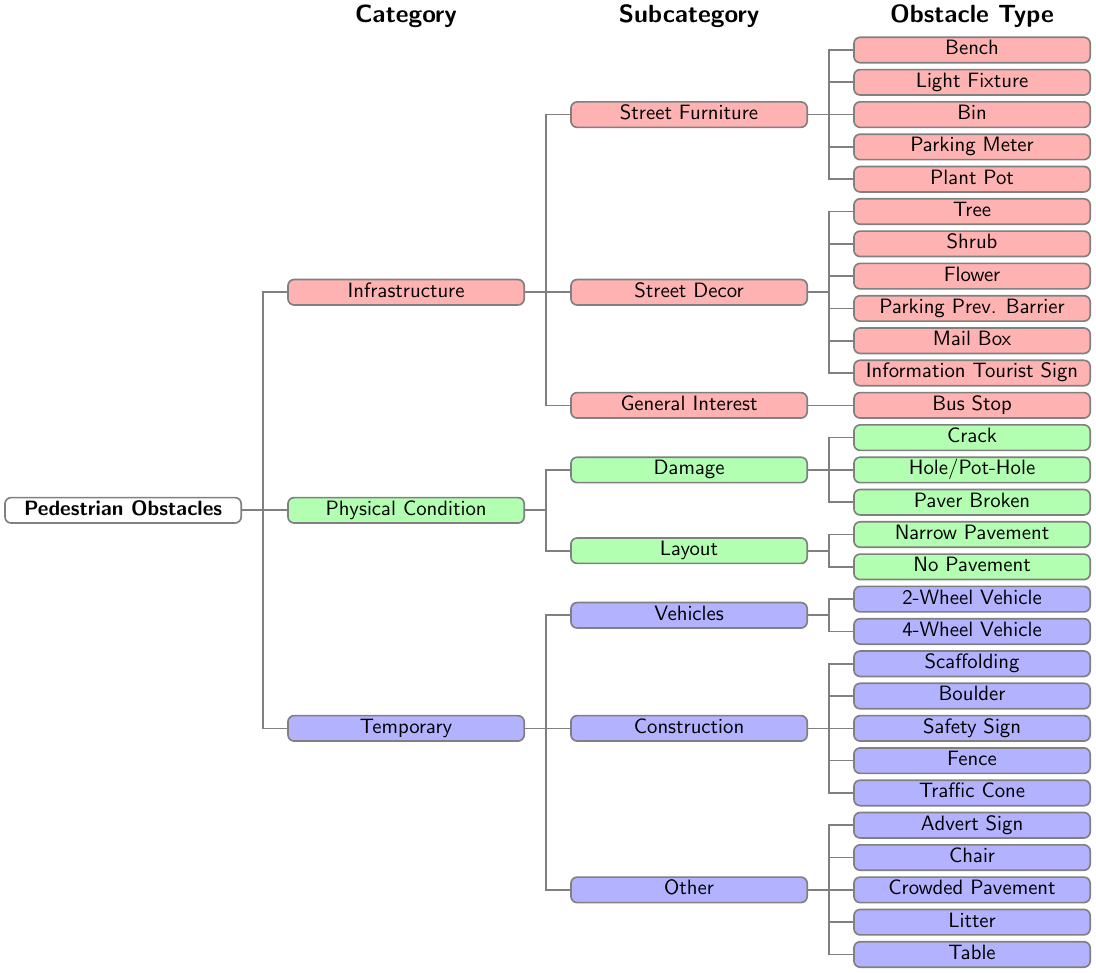}
    \caption{The taxonomy of obstacle types included in the dataset.}
    \label{fig:taxonomy}
\end{figure}

In this work, we aim to create and make publicly available a new comprehensive visual dataset related to pedestrian safety, to address a gap in the related literature.
In this section, we describe the methodology we followed to create the new \pedestrian dataset, and give an overview of the created dataset.
Detailed instructions on how to obtain the dataset are included in Appendix~\ref{appendix:dataset-availability}.

\subsection{Data Collection}

The \pedestrian dataset is composed of video files showcasing obstacles that endanger the safety of pedestrians when walking in the city of Nicosia, Cyprus.
Proper identification of obstacles requires special attention to various considerations: defining the different obstacle types, assessing the degree of danger they pose, and determining in what context a simple object can be considered an obstacle.
Thus, the problems related to pedestrian safety in modern cities were studied, and the various kinds of obstacles observed on city sidewalks were identified before collecting the data.
In total, 29 different types of obstacles (see Figure~\ref{fig:dataset-image-examples}) were identified and used for data collection.

Obstacles were grouped into 3 high-level categories:
\begin{enumerate*}[(i)]
    \item \taxLevel{Physical Condition},
    \item \taxLevel{Infrastructure}, and
    \item \taxLevel{Temporary},
\end{enumerate*}
as illustrated in Figure~\ref{fig:taxonomy}.
The \taxLevel{Physical Condition} category includes obstacles related to poor pavement maintenance, which is further divided into 2 subcategories.
The \taxLevel{Infrastructure} category includes obstacles related to parts of the infrastructure that affect safe traffic on a sidewalk or part of the sidewalk and consists of 3 subcategories.
Finally, the \taxLevel{Temporary} category concerns all temporary obstacles that affect the safety of sidewalks and concerns 3 subcategories.

The data collection process was carefully designed to guarantee a dataset that is both comprehensive and varied, particularly in terms of natural lighting conditions.
We compiled the dataset by recording videos, each focusing on a distinct obstacle type.
Subsequently, we extracted individual frames from these videos to create the dataset, as illustrated in Figure~\ref{fig:frame-extraction}.
This approach ensured that each video precisely showcased a single obstacle, enabling us to efficiently capture the wide range of barriers encountered in urban environments with a high degree of control and accuracy.

The data collection process was facilitated by the use of a simple and low-cost setup, comprising of smartphone cameras.
Two smartphone models were used, in order to prevent DL models overfitting on the characteristics of a single capture device.
Specifically, the cameras of the smartphones Xiaomi Mi~Mix~3 and Apple iPhone~7 were used.
The videos from both smartphone cameras were captured at chest height, simulating the point of view of a wearable camera worn by a pedestrian.
The dataset consists of 340 videos, covering the 29 predefined obstacle types, under 3 different lighting conditions: daytime with sunlight, cloudy day, and night.
Data collection was followed by a processing step to resolve privacy issues and comply with the EU General Data Protection Regulation (GDPR)~\cite{eu2016GDPR}, which included face, car license plate, and signs blurring.

\subsection{Dataset Description}

A breakdown of the dataset statistics for the 3 taxonomy levels, and a breakdown across lighting conditions are shown in Tables~\ref{tab:summary_table_category},~\ref{tab:summary_table_subcategory}, and~\ref{tab:summary_table_obstacle_type}, respectively.

\begin{figure*}[t]
    \centering
    \subfloat[\centering Number of videos per obstacle type]{{\includegraphics[width=0.45\textwidth]{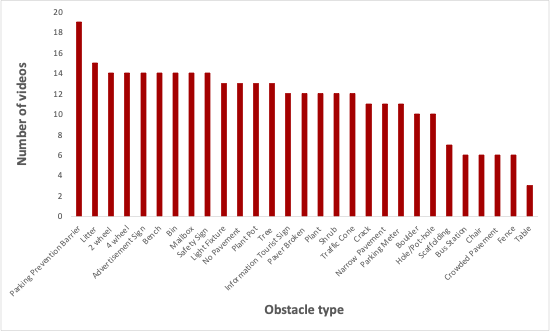} }\label{fig2a}}%
    \qquad
    \subfloat[\centering Histogram of video duration]{{\includegraphics[width=0.45\textwidth]{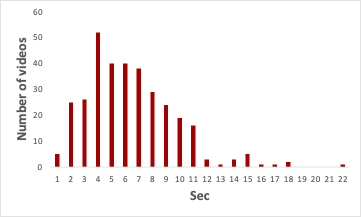} }\label{fig2c}}%
    \caption{Statistics of the dataset.}%
    \label{fig2}
\end{figure*}

\begin{figure}[t]
    \centering
    \includegraphics[width=1\textwidth]{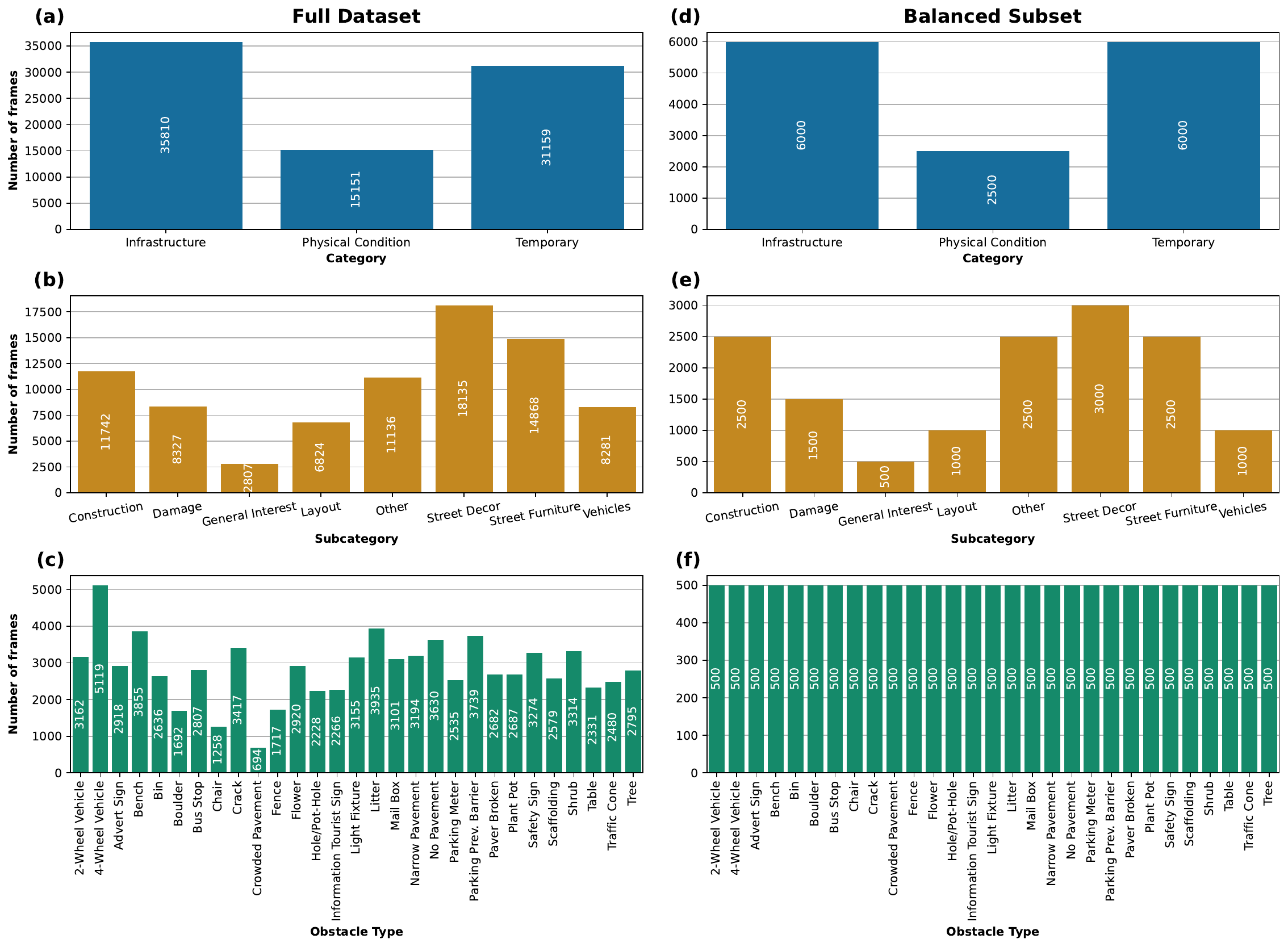}
    \caption{
        Breakdown of the number of frames in the full \pedestrian dataset for the three taxonomy levels: (a) \taxLevel{Category}, (b) \taxLevel{Subcategory}, and (c) \taxLevel{Obstacle type}; and, respectively, (d--f) for the balanced subset used in the benchmarking experiments.
    }
    \label{fig:dataset-stats}
\end{figure}

The total number of videos per \taxLevel{Obstacle type} are presented in Figure~\ref{fig2a}.
Figure~\ref{fig2c} shows a histogram of video duration, which varies from 1s to 17s.
Overall, the dataset consists of a total of $82120$ images.
Figure~\ref{fig:dataset-stats}a--c shows a visual breakdown of the number of images per taxonomy level.

All videos in the dataset have a Full~HD resolution of $1080\times1920$ pixels (portrait orientation).
The default settings for video frame rates (frames-per-second, FPS) for the two smartphones were used: 60~FPS for the Mi~Mix~3 and 30~FPS for the iPhone~7.
Table~\ref{tab:summary_table_camera_model_fps} shows of a breakdown of the dataset statistics based on smartphone model (and by proxy, FPS).

\begin{table*}[t]
    \centering
    \caption{Number of videos, total video length and available frames per \taxLevel{Category} and lighting condition.}
    \label{tab:summary_table_category}
    \resizebox{\textwidth}{!}{
        \begin{tabular}{lrrrrrrrrrrrr}
            \toprule
                                              & \multicolumn{4}{c}{Video Count} & \multicolumn{4}{c}{Total Video Length (s)} & \multicolumn{4}{c}{Frame Count}                                                                         \\ \cmidrule(r){2-5} \cmidrule(r){6-9} \cmidrule(r){10-13}
            \multirow{-2}{*}[0.5ex]{Category} & Day                             & Night                                      & Cloudy                          & Total & Day & Night & Cloudy & Total & Day   & Night & Cloudy & Total \\
            \midrule
            Infrastructure                    & 53                              & 53                                         & 47                              & 153   & 361 & 346   & 321    & 1029  & 10832 & 14863 & 10115  & 35810 \\
            Physical Condition                & 20                              & 19                                         & 20                              & 59    & 128 & 138   & 150    & 417   & 3867  & 6777  & 4507   & 15151 \\
            Temporary                         & 44                              & 44                                         & 40                              & 128   & 339 & 299   & 246    & 885   & 10185 & 11984 & 8990   & 31159 \\
            \midrule
            Total                             & 117                             & 116                                        & 107                             & 340   & 829 & 784   & 718    & 2332  & 24884 & 33624 & 23612  & 82120 \\
            \bottomrule
        \end{tabular}
    }
\end{table*}

\begin{table*}[t]
    \centering
    \caption{Number of videos, total video length (in seconds) and available frames per \taxLevel{Subcategory} and lighting condition.}
    \label{tab:summary_table_subcategory}
    \resizebox{\textwidth}{!}{
        \begin{tabular}{lrrrrrrrrrrrr}
            \toprule
                                                 & \multicolumn{4}{c}{Video Count} & \multicolumn{4}{c}{Total Video Length (s)} & \multicolumn{4}{c}{Frame Count}                                                                         \\ \cmidrule(r){2-5} \cmidrule(r){6-9} \cmidrule(r){10-13}
            \multirow{-2}{*}[0.5ex]{Subcategory} & Day                             & Night                                      & Cloudy                          & Total & Day & Night & Cloudy & Total & Day   & Night & Cloudy & Total \\
            \midrule
            Construction                         & 16                              & 16                                         & 17                              & 49    & 149 & 109   & 103    & 361   & 4485  & 3716  & 3541   & 11742 \\
            Damage                               & 13                              & 11                                         & 11                              & 35    & 78  & 79    & 62     & 220   & 2367  & 4082  & 1878   & 8327  \\
            General Interest                     & 2                               & 2                                          & 2                               & 6     & 13  & 26    & 26     & 66    & 400   & 1602  & 805    & 2807  \\
            Layout                               & 7                               & 8                                          & 9                               & 24    & 50  & 59    & 87     & 197   & 1500  & 2695  & 2629   & 6824  \\
            Other                                & 18                              & 19                                         & 16                              & 53    & 116 & 118   & 80     & 315   & 3498  & 4752  & 2886   & 11136 \\
            Street Decor                         & 27                              & 29                                         & 26                              & 82    & 183 & 189   & 168    & 542   & 5509  & 7407  & 5219   & 18135 \\
            Street Furniture                     & 24                              & 22                                         & 19                              & 65    & 164 & 130   & 125    & 420   & 4923  & 5854  & 4091   & 14868 \\
            Vehicles                             & 10                              & 9                                          & 7                               & 26    & 73  & 72    & 63     & 208   & 2202  & 3516  & 2563   & 8281  \\
            \midrule
            Total                                & 117                             & 116                                        & 107                             & 340   & 829 & 784   & 718    & 2332  & 24884 & 33624 & 23612  & 82120 \\
            \bottomrule
        \end{tabular}
    }
\end{table*}

\begin{table*}[t]
    \centering
    \caption{Number of videos, total video length (in seconds) and available frames per \taxLevel{Obstacle type} and lighting condition.}
    \label{tab:summary_table_obstacle_type}
    \resizebox{\textwidth}{!}{
        \begin{tabular}{lrrrrrrrrrrrr}
            \toprule
                                                   & \multicolumn{4}{c}{Video Count} & \multicolumn{4}{c}{Total Video Length (s)} & \multicolumn{4}{c}{Frame Count}                                                                         \\ \cmidrule(r){2-5} \cmidrule(r){6-9} \cmidrule(r){10-13}
            \multirow{-2}{*}[0.5ex]{Obstacle type} & Day                             & Night                                      & Cloudy                          & Total & Day & Night & Cloudy & Total & Day   & Night & Cloudy & Total \\
            \midrule
            2-Wheel Vehicle                        & 6                               & 5                                          & 3                               & 14    & 34  & 25    & 20     & 79    & 1022  & 1533  & 607    & 3162  \\
            4-Wheel Vehicle                        & 4                               & 4                                          & 4                               & 12    & 39  & 46    & 42     & 128   & 1180  & 1983  & 1956   & 5119  \\
            Advert Sign                            & 6                               & 4                                          & 4                               & 14    & 45  & 27    & 13     & 86    & 1362  & 1146  & 410    & 2918  \\
            Bench                                  & 6                               & 4                                          & 4                               & 14    & 43  & 24    & 35     & 103   & 1309  & 1487  & 1059   & 3855  \\
            Bin                                    & 6                               & 5                                          & 3                               & 14    & 37  & 23    & 14     & 75    & 1138  & 1070  & 428    & 2636  \\
            Boulder                                & 2                               & 4                                          & 4                               & 10    & 9   & 28    & 18     & 56    & 296   & 845   & 551    & 1692  \\
            Bus Stop                               & 2                               & 2                                          & 2                               & 6     & 13  & 26    & 26     & 66    & 400   & 1602  & 805    & 2807  \\
            Chair                                  & 2                               & 2                                          & 2                               & 6     & 13  & 10    & 7      & 31    & 410   & 622   & 226    & 1258  \\
            Crack                                  & 4                               & 4                                          & 3                               & 11    & 23  & 31    & 26     & 82    & 715   & 1902  & 800    & 3417  \\
            Crowded Pavement                       & 2                               & 2                                          & 2                               & 6     & 8   & 4     & 7      & 20    & 259   & 213   & 222    & 694   \\
            Fence                                  & 2                               & 2                                          & 2                               & 6     & 20  & 18    & 18     & 57    & 617   & 551   & 549    & 1717  \\
            Flower                                 & 2                               & 6                                          & 4                               & 12    & 16  & 35    & 34     & 85    & 485   & 1410  & 1025   & 2920  \\
            Hole/Pot-Hole                          & 4                               & 4                                          & 4                               & 12    & 22  & 23    & 16     & 63    & 686   & 1045  & 497    & 2228  \\
            Information Tourist Sign               & 4                               & 4                                          & 4                               & 12    & 28  & 26    & 20     & 75    & 851   & 802   & 613    & 2266  \\
            Light Fixture                          & 4                               & 4                                          & 5                               & 13    & 23  & 24    & 36     & 84    & 700   & 1059  & 1396   & 3155  \\
            Litter                                 & 4                               & 7                                          & 4                               & 15    & 29  & 54    & 24     & 107   & 874   & 1867  & 1194   & 3935  \\
            Mail Box                               & 4                               & 6                                          & 4                               & 14    & 20  & 46    & 23     & 90    & 613   & 1791  & 697    & 3101  \\
            Narrow Pavement                        & 3                               & 4                                          & 4                               & 11    & 23  & 30    & 38     & 92    & 708   & 1343  & 1143   & 3194  \\
            No Pavement                            & 4                               & 4                                          & 5                               & 13    & 26  & 28    & 49     & 104   & 792   & 1352  & 1486   & 3630  \\
            Parking Meter                          & 4                               & 4                                          & 3                               & 11    & 30  & 23    & 21     & 75    & 926   & 963   & 646    & 2535  \\
            Parking Prev. Barrier                  & 6                               & 6                                          & 7                               & 19    & 35  & 33    & 40     & 109   & 1069  & 1305  & 1365   & 3739  \\
            Paver Broken                           & 5                               & 3                                          & 4                               & 12    & 32  & 23    & 19     & 75    & 966   & 1135  & 581    & 2682  \\
            Plant Pot                              & 4                               & 5                                          & 4                               & 13    & 28  & 33    & 18     & 81    & 850   & 1275  & 562    & 2687  \\
            Safety Sign                            & 4                               & 4                                          & 6                               & 14    & 28  & 20    & 36     & 86    & 870   & 871   & 1533   & 3274  \\
            Scaffolding                            & 4                               & 2                                          & 1                               & 7     & 55  & 22    & 7      & 85    & 1670  & 670   & 239    & 2579  \\
            Shrub                                  & 6                               & 3                                          & 3                               & 12    & 53  & 23    & 20     & 97    & 1590  & 1106  & 618    & 3314  \\
            Table                                  & 4                               & 4                                          & 4                               & 12    & 19  & 20    & 27     & 68    & 593   & 904   & 834    & 2331  \\
            Traffic Cone                           & 4                               & 4                                          & 4                               & 12    & 34  & 19    & 22     & 75    & 1032  & 779   & 669    & 2480  \\
            Tree                                   & 5                               & 4                                          & 4                               & 13    & 30  & 23    & 30     & 83    & 901   & 993   & 901    & 2795  \\
            \midrule
            Total                                  & 117                             & 116                                        & 107                             & 340   & 829 & 784   & 718    & 2332  & 24884 & 33624 & 23612  & 82120 \\
            \bottomrule
        \end{tabular}
    }
\end{table*}

\begin{table*}[t]
    \centering
    \caption{Number of videos, total video length (in seconds) and available frames per smartphone model (and by extension, frame rate) and lighting condition.}
    \label{tab:summary_table_camera_model_fps}
    \resizebox{\textwidth}{!}{
        \begin{tabular}{lrrrrrrrrrrrr}
            \toprule
                                                  & \multicolumn{4}{c}{Video Count} & \multicolumn{4}{c}{Total Video Length (s)} & \multicolumn{4}{c}{Frame Count}                                                                         \\ \cmidrule(r){2-5} \cmidrule(r){6-9} \cmidrule(r){10-13}
            \multirow{-2}{*}[0.5ex]{Camera model} & Day                             & Night                                      & Cloudy                          & Total & Day & Night & Cloudy & Total & Day   & Night & Cloudy & Total \\
            \midrule
            iPhone 7 (30 FPS)                     & 117                             & 60                                         & 98                              & 275   & 829 & 448   & 650    & 1927  & 24884 & 13445 & 19508  & 57837 \\
            Mi Mix 3 (60 FPS)                     & 0                               & 56                                         & 9                               & 65    & 0   & 336   & 68     & 405   & 0     & 20179 & 4104   & 24283 \\
            \midrule
            Total                                 & 117                             & 116                                        & 107                             & 340   & 829 & 784   & 718    & 2332  & 24884 & 33624 & 23612  & 82120 \\
            \bottomrule
        \end{tabular}
    }
\end{table*}

\section{Benchmark Experiments} \label{sec:benchmarks}

A diverse set of benchmarking experiments was performed to evaluate the utility of the new \pedestrian dataset presented in this work.
Specifically, a number of pretrained DL model architectures were fine-tuned on a balanced subset of the dataset, to evaluate its performance.
See Appendix~\ref{appendix:benchmark-experiments} for details.

\subsection{Balanced Dataset Subset} \label{sec:balanced-dataset-subset}

\begin{table*}[t]
    \centering
    \caption{Model architectures used in the fine-tuning experiments.
        For all models ImageNet-1K pretrained weights provided by PyTorch were used (links to the weights and model recipes are provided).
        Accuracies are reported on the ImageNet-1K dataset using single crops.
        The models are sorted in ascending order in regards to their number of parameters (in millions).
    }
    \label{tab:models}
    \begin{tabular}{lrrrccc}
        \toprule
        Model Name                & \thead{Top-1                                                                                                                                                                                                                                                         \\Acc. (\%)} & \thead{Top-5\\Acc. (\%)} & \thead{Params\\(M) $\uparrow$} & Weights & \thead{Model\\Recipe} & Citation \\
        \midrule
        \model{MobileNetV3-Small} & 67.67        & 87.40 & 2.5   & \href{https://download.pytorch.org/models/mobilenet_v3_small-047dcff4.pth}{link}        & \href{https://github.com/pytorch/vision/tree/main/references/classification#mobilenetv3-large--small}{link} & \cite{howard2019MobileNetV3}  \\
        \model{MobileNetV2}       & 71.88        & 90.29 & 3.5   & \href{https://download.pytorch.org/models/mobilenet_v2-b0353104.pth}{link}              & \href{https://github.com/pytorch/vision/tree/main/references/classification#mobilenetv2}{link}              & \cite{sandler2018MobileNetV2} \\
        \model{EfficientNet-B0}   & 77.69        & 93.53 & 5.3   & \href{https://download.pytorch.org/models/efficientnet_b0_rwightman-7f5810bc.pth}{link} & \href{https://github.com/pytorch/vision/tree/main/references/classification#efficientnet-v1}{link}          & \cite{tan2020EfficientNet}    \\
        \model{MobileNetV3-Large} & 74.04        & 91.34 & 5.5   & \href{https://download.pytorch.org/models/mobilenet_v3_large-8738ca79.pth}{link}        & \href{https://github.com/pytorch/vision/tree/main/references/classification#mobilenetv3-large--small}{link} & \cite{howard2019MobileNetV3}  \\
        \model{GoogLeNet}         & 69.78        & 89.53 & 6.6   & \href{https://download.pytorch.org/models/googlenet-1378be20.pth}{link}                 & \href{https://github.com/pytorch/vision/tree/main/references/classification#googlenet}{link}                & \cite{szegedy2014GoogLeNet}   \\
        \model{DenseNet-121}      & 74.43        & 91.97 & 8.0   & \href{https://download.pytorch.org/models/densenet121-a639ec97.pth}{link}               & \href{https://github.com/pytorch/vision/pull/116}{link}                                                     & \cite{huang2017DenseNet}      \\
        \model{ResNet-18}         & 69.76        & 89.08 & 11.7  & \href{https://download.pytorch.org/models/resnet18-f37072fd.pth}{link}                  & \href{https://github.com/pytorch/vision/tree/main/references/classification#resnet}{link}                   & \cite{he2015DeepResidual}     \\
        \model{DenseNet-201}      & 76.90        & 93.37 & 20.0  & \href{https://download.pytorch.org/models/densenet201-c1103571.pth}{link}               & \href{https://github.com/pytorch/vision/pull/116}{link}                                                     & \cite{huang2017DenseNet}      \\
        \model{EfficientNetV2-S}  & 84.23        & 96.88 & 21.5  & \href{https://download.pytorch.org/models/efficientnet_v2_s-dd5fe13b.pth}{link}         & \href{https://github.com/pytorch/vision/tree/main/references/classification#efficientnet-v2}{link}          & \cite{tan2021EfficientNetV2}  \\
        \model{ResNet-34}         & 73.31        & 91.42 & 21.8  & \href{https://download.pytorch.org/models/resnet34-b627a593.pth}{link}                  & \href{https://github.com/pytorch/vision/tree/main/references/classification#resnet}{link}                   & \cite{he2015DeepResidual}     \\
        \model{ResNet-50}         & 76.13        & 92.86 & 25.6  & \href{https://download.pytorch.org/models/resnet50-0676ba61.pth}{link}                  & \href{https://github.com/pytorch/vision/tree/main/references/classification#resnet}{link}                   & \cite{he2015DeepResidual}     \\
        \model{ResNet-101}        & 77.37        & 93.55 & 44.5  & \href{https://download.pytorch.org/models/resnet101-63fe2227.pth}{link}                 & \href{https://github.com/pytorch/vision/tree/main/references/classification#resnet}{link}                   & \cite{he2015DeepResidual}     \\
        \model{ConvNeXt-Small}    & 83.62        & 96.65 & 50.2  & \href{https://download.pytorch.org/models/convnext_small-0c510722.pth}{link}            & \href{https://github.com/pytorch/vision/tree/main/references/classification#convnext}{link}                 & \cite{liu2022ConvNeXt}        \\
        \model{ResNet-152}        & 78.31        & 94.05 & 60.2  & \href{https://download.pytorch.org/models/resnet152-394f9c45.pth}{link}                 & \href{https://github.com/pytorch/vision/tree/main/references/classification#resnet}{link}                   & \cite{he2015DeepResidual}     \\
        \model{ConvNeXt-Base}     & 84.06        & 96.87 & 88.6  & \href{https://download.pytorch.org/models/convnext_base-6075fbad.pth}{link}             & \href{https://github.com/pytorch/vision/tree/main/references/classification#convnext}{link}                 & \cite{liu2022ConvNeXt}        \\
        \model{ConvNeXt-Large}    & 84.41        & 96.98 & 197.8 & \href{https://download.pytorch.org/models/convnext_large-ea097f82.pth}{link}            & \href{https://github.com/pytorch/vision/tree/main/references/classification#convnext}{link}                 & \cite{liu2022ConvNeXt}        \\
        \bottomrule
    \end{tabular}
\end{table*}

To facilitate the benchmark experiments, and to ensure a balanced dataset in terms of images per \taxLevel{Obstacle type}, we extracted a subset from the \pedestrian dataset described in the previous sections, using the pipeline shown in Figure~\ref{fig:frame-extraction}.
For each obstacle type, a fixed number of images $c$ was extracted from the total number of available frames per obstacle type, $n$, ensuring that $c \leq n$ for all types.
Given that each obstacle type had a different number of frames available (refer to Table~\ref{tab:summary_table_obstacle_type}), the frames were selected as follows:
\[
    \text{frame selection}(n, c) =
    \begin{cases}
        \text{uniformly at random},                                             & \text{if } 1 \leq \frac{n}{c} < 2, \\
        \text{regularly at intervals of } \left\lfloor\frac{n}{c}\right\rfloor, & \text{if } \frac{n}{c} \geq 2.
    \end{cases}
\]
For the regular interval selection, the selection process started from the first frame available (index 0 of the concatenated frames from all videos for each obstacle type),  and continued at intervals of \(\left\lfloor\frac{n}{c}\right\rfloor\).

To create the balanced subset, $c=500$ was chosen, that accommodates the minimum available number of frames for the obstacle type \taxLevel{Crowded Pavement} ($n=694$, see Table~\ref{tab:summary_table_obstacle_type}).
Thus, the size of the balanced subset of \pedestrian that was used for the fine-tuning experiments was $29 \times 500 = 14500$ images.
Figure~\ref{fig:dataset-stats}d--f shows a visual breakdown of the number of images per taxonomy level in the balanced subset (note that while that the balancing was done only in terms of images per \taxLevel{Obstacle type}, and not for the other two taxonomy levels).

\subsection{Model Fine-tuning}

A total of 16 DL architectures, listed in Table~\ref{tab:models}, were fine-tuned using the balanced subset described in the previous section.
Effort was made to include state-of-the art model architectures, especially in regards to parameter size: the smallest model used in terms of parameters was \model{MobileNetV3-Small} at 2.5M parameters~\cite{howard2019MobileNetV3}, while the largest was \model{ConvNeXt-Large} at 197.8M parameters~\cite{liu2022ConvNeXt}.
The pretrained weights of the models using the ImageNet dataset~\cite{deng2009ImageNet,russakovsky2015ImageNet}, that are available through the PyTorch library (links provided in Table~\ref{tab:models}), were used.

Each of the architectures was fine-tuned to identify the obstacles at all three taxonomy levels (\taxLevel{Category}, \taxLevel{Subcategory} and \taxLevel{Obstacle type}).
In addition, the models were trained by either freezing all their layers apart from the last, or without freezing any of their layers.
A random split of the dataset into 70\% training, 20\% validation and 10\% testing subsets was performed before training.
Training was performed for 30 epochs, with batches of 32 images, using the Adam optimizer~\cite{kingma2017AdamOptimizer} and a learning rate of $0.001$.
The default pre-processing transforms provided by the model recipes (see Table~\ref{tab:models}) were applied to the input images during training and testing.

For each training combination of DL architecture and taxonomy level, the experiments were repeated 5 times, using a different random seed for the splitting into training/validation/testing subsets.
The same sequence of random seeds was used for each combination, to enable one-to-one comparisons between combinations.
Thus, a total of $16\text{ DL architectures} \times 3 \text{ tax. levels} \times 2 \text{ layer freezing status} \times 5 \text{ reps} = 480$ fine-tuning experiments were performed.

\subsection{Experimental Results}

\begin{figure}[t]
    \centering
    \includegraphics[width=\textwidth]{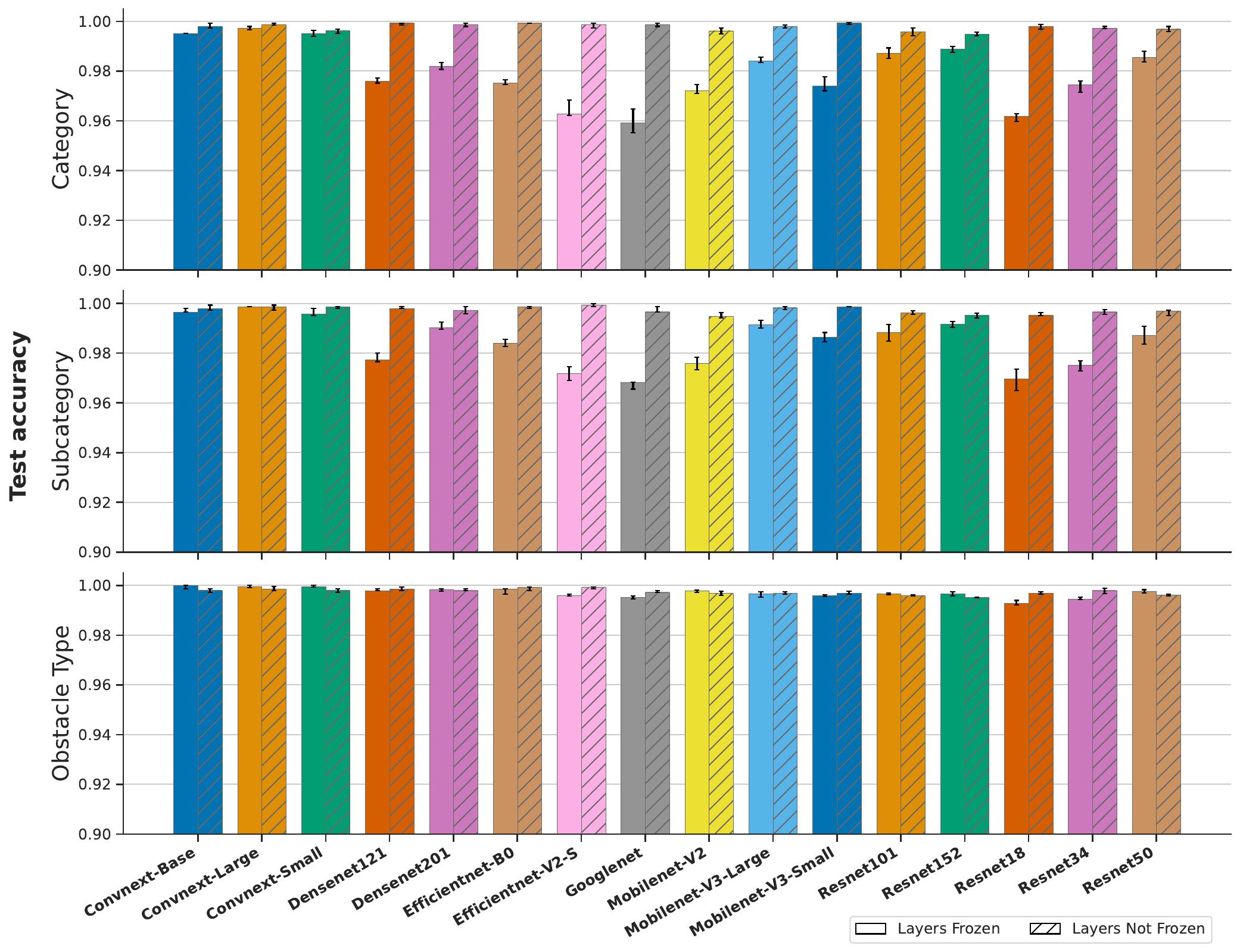}
    \caption{
        Benchmark results for the fine-tuning of the 16 DL models using the \pedestrian dataset, on all 3 taxonomy levels, \textit{with} model layer freezing (bars with no hatches) or \textit{without} layer freezing (bars with hatches).
        The fine-tuning took place over 30 epochs; the median test accuracy results for the best epoch of each training combination are shown, over 5 repetitions that were run for each combination.
        The error bars show the Q1--Q3 interval.
        The y-axis scale is truncated to range $[0.9,1.0]$ to make the differences between models more immediately obvious.}
    \label{fig:benchmark_results}
\end{figure}

\begin{table*}[t]
    \centering
    \caption{Benchmark results for the fine-tuning of the 16 DL models using the \pedestrian dataset, on all 3 taxonomy levels, either with frozen (F) or not frozen (NF) model layers.
        The mean test accuracy results for the best epoch of each training combination are shown, averaged over the 5 repetitions that were run for each combination.
        The min and max values for each column are highlighted in \textit{italics} and \textbf{bold}, respectively.}
    \label{tab:benchmark_results}
    \resizebox{\textwidth}{!}{
        \begin{tabular}{lrrrrrr}
            \toprule
                                                & \multicolumn{2}{c}{Category} & \multicolumn{2}{c}{Subcategory} & \multicolumn{2}{c}{Obstacle Type}                                                       \\ \cmidrule(r){2-3} \cmidrule(r){4-5} \cmidrule(r){6-7}
            \multirow{-2}{*}[0.5ex]{Model Name} & NF                           & F                               & NF                                & F               & NF              & F               \\
            \midrule
            \model{ConvNeXt-Small}              & 99.586                       & 99.509                          & 99.793                            & 99.612          & 99.747          & \textbf{99.959} \\
            \model{ConvNeXt-Base}               & 99.793                       & 99.545                          & 99.821                            & 99.738          & 99.807          & 99.931          \\
            \model{ConvNeXt-Large}              & 99.848                       & \textbf{99.614}                 & 99.807                            & \textbf{99.862} & 99.879          & 99.948          \\
            \model{DenseNet-121}                & 99.862                       & 97.626                          & 99.809                            & 97.810          & 99.847          & 99.801          \\
            \model{DenseNet-201}                & 99.848                       & 98.208                          & 99.711                            & 99.076          & 99.779          & 99.793          \\
            \model{EfficientNet-B0}             & \textbf{99.904}              & 97.491                          & 99.821                            & 98.346          & 99.876          & 99.793          \\
            \model{EfficientNetV2-S}            & 99.821                       & 96.485                          & \textbf{99.917}                   & 97.161          & \textbf{99.890} & 99.614          \\
            \model{GoogLeNet}                   & 99.835                       & \textit{95.879}                 & 99.724                            & \textit{96.637} & 99.766          & 99.531          \\
            \model{MobileNetV2}                 & 99.621                       & 97.364                          & 99.552                            & 97.588          & 99.673          & 99.742          \\
            \model{MobileNetV3-Large}           & 99.793                       & 98.501                          & 99.793                            & 99.190          & 99.690          & 99.604          \\
            \model{MobileNetV3-Small}           & 99.897                       & 97.553                          & 99.879                            & 98.622          & 99.707          & 99.586          \\
            \model{ResNet-18}                   & 99.776                       & 96.089                          & 99.604                            & 96.881          & 99.707          & \textit{99.328} \\
            \model{ResNet-34}                   & 99.776                       & 97.295                          & 99.655                            & 97.467          & 99.776          & 99.518          \\
            \model{ResNet-50}                   & 99.690                       & 98.622                          & 99.552                            & 98.725          & 99.604          & 99.759          \\
            \model{ResNet-101}                  & 99.586                       & 98.725                          & 99.638                            & 98.811          & 99.604          & 99.655          \\
            \model{ResNet-152}                  & \textit{99.500}              & 98.828                          & \textit{99.500}                   & 99.139          & \textit{99.518} & 99.655          \\
            \bottomrule
        \end{tabular}
    }
\end{table*}
The results obtained from the fine-tuning of the 16 DL models utilizing the \pedestrian dataset at all three taxonomy levels are summarized in Figure~\ref{fig:benchmark_results} and Table~\ref{tab:benchmark_results}.
(Detailed results, broken down per \taxLevel{Obstacle type}, are provided in Appendix~\ref{appendix:detailed-results}, Tables~\ref{tab:acc-obstacle-type-detailed-part1}~\&~\ref{tab:acc-obstacle-type-detailed-part2}).
These results include the outcomes from experiments both with and without freezing the layers of the models.
The primary objective of these experiments was to assess the effectiveness of the new \pedestrian dataset across a diverse range of DL architectures.

At the \taxLevel{Obstacle type} level, all models produced very good results, with accuracies over 99\%.
\model{ConvNeXt-Small}~\cite{liu2022ConvNeXt} had a slight advantage with 99.96\% accuracy with its layers frozen, while \model{EfficientNetV2-S}~\cite{tan2021EfficientNetV2} led with 99.89\% accuracy without layer freezing.
On the lower end, \model{ResNet-152} and \model{ResNet-18}~\cite{he2015DeepResidual} achieved 99.52\% and 99.33\% respectively, the former with unfrozen layers and the latter with layers frozen.

For the \taxLevel{Obstacle type} level, the impact of freezing the layers on model accuracy was negligible.
In contrast, for the \taxLevel{Category} and \taxLevel{Subcategory} levels, allowing the layers to remain unfrozen generally yielded better results.
Specifically, at the \taxLevel{Category} level, \model{EfficientNet-B0}~\cite{tan2020EfficientNet} achieved the highest accuracy of 99.90\% with unfrozen layers, while \model{ConvNeXt-Large}~\cite{liu2022ConvNeXt} topped the frozen layers category with 99.61\% accuracy.
For the \taxLevel{Subcategory} level, \model{EfficientNetV2-S} reached the highest unfrozen accuracy at~99.92\%, and \model{ConvNeXt-Large} again performed best for the models with frozen layers at~99.86\%.

The observed discrepancy in model performance between the \taxLevel{Category} and \taxLevel{Subcategory} levels, as compared to the \taxLevel{Obstacle type} level, may stem from several factors.
It was noted that networks with fewer parameters were, in general, less adept at handling the complexity required for \taxLevel{Category} and \taxLevel{Subcategory} classification tasks.
This struggle may be partially attributed to an imbalance within the dataset for these taxonomy levels, since even though the subset used was balanced in terms of number of frames for the taxonomy level \taxLevel{Obstacle type} (Figure~\ref{fig:dataset-stats}f), it was not balanced for the other two levels (Figure~\ref{fig:dataset-stats}d~\&~\ref{fig:dataset-stats}e).
Furthermore, the difference in performance can potentially be explained by the necessity for the networks to abstract and learn more complex features to accurately classify data at the \taxLevel{Category} and \taxLevel{Subcategory} levels.
These factors suggest that the \taxLevel{Category} and \taxLevel{Subcategory} levels present inherently more challenging classification problems, revealing a discernible discrepancy in model efficacy across taxonomy levels.

The results gathered from the series of experiments were, in general, excellent.
The performance metrics indicated that the \pedestrian dataset is robust and enables the DL models to achieve high levels of accuracy.
Across different architectures, there was a consistent trend of models being able to effectively recognize and classify the data, regardless of the taxonomic level considered.
This underlines the \pedestrian dataset's quality and its utility in training DL models for pedestrian obstacle detection and classification tasks.

\section{Conclusions} \label{sec:conclusion}

This work addresses the critical issue of pedestrian safety, which remains a significant concern globally, due to the risks posed by road accidents.
Recognizing the lack of specialized resources for enhancing pedestrian safety through obstacle detection and avoidance, we introduced the \pedestrian dataset, a novel egocentric vision dataset specifically designed for identifying obstacles that pedestrians commonly encounter in modern urban environments.
The dataset comprises of 340 videos, collected and processed to represent 29 distinct types of obstacles under various lighting conditions, using smartphone cameras to simulate a pedestrian's point of view.
This effort marks a significant step towards leveraging advanced machine learning techniques, particularly deep learning, to improve pedestrian safety by providing a comprehensive dataset that captures the complexity and diversity of urban pedestrian obstacles.

To demonstrate the utility and effectiveness of the \pedestrian dataset, we conducted a series of benchmark experiments involving the fine-tuning of 16 different deep learning architectures on the dataset.
These experiments were designed to evaluate the dataset's performance across three taxonomic levels of obstacle classification: \taxLevel{Category}, \taxLevel{Subcategory} and \taxLevel{Obstacle type}.
By performing fine-tuning with both frozen and unfrozen model layers, we aimed to explore the adaptability and learning capabilities of various architectures when applied to the task of pedestrian obstacle detection.
The benchmark experiments served a dual purpose: firstly, to validate the \pedestrian dataset as a robust tool for training and enhancing obstacle detection models, and secondly, to provide insights into the effectiveness of different deep learning architectures for egocentric vision tasks in urban pedestrian settings.
Through these experiments, we established a comprehensive evaluation framework that underscores the dataset's potential to advance research and development in pedestrian safety technologies.

The \pedestrian dataset has great potential for developing effective applications and systems to enhance the safety of citizens as they navigate city sidewalks, especially given the increased distractions from mobile phone use.
Additionally, this dataset can aid in creating systems that assist individuals using mobility aids, including those with mobility impairments and the elderly, ensuring their comfortable and safe movement on urban sidewalks.

In future work, several directions could be pursued to enhance the capabilities and applications of pedestrian safety technologies.
Expanding the \pedestrian dataset to cover a wider array of environments and conditions would be beneficial for improving the robustness of obstacle detection models.
The collection of images from other cities/countries is one of the main future goals for expanding this work.
Synergies with partners in other countries will be explored to enrich the dataset.
Incorporating additional data types, such as audio and sensor data, could also enrich the dataset and enable more comprehensive obstacle detection solutions.
Moreover, investigating the use of federated learning could facilitate the collaborative enhancement of models while addressing privacy concerns.
Collaboration with urban planners could also provide insights into practical implementations of these technologies.
These future research directions underscore the importance of continued exploration and interdisciplinary collaboration to advance pedestrian safety technologies.

In conclusion, this work presents the \pedestrian dataset, a new egocentric dataset focused on pedestrian safety.
A suite of benchmark experiments using different deep learning models show the dataset's usefulness for the development of effective obstacle detection systems.
The main importance of this work is its contribution to the goal of making pedestrian environments safer, setting the stage for future advancements in this essential field.

\backmatter

\section*{Declarations}

\bmhead{Funding}

This project has received funding from the European Union's Horizon 2020 Research and Innovation Programme, under Grant Agreement No 739578, complemented by the Government of the Republic of Cyprus through the Directorate General for European Programmes, Coordination and Development.

\bmhead{Conflict of Interest}

On behalf of all authors, the corresponding author states that there is no conflict of interest.



\pagebreak

\begin{appendices}

    \section{Dataset \& Benchmarks}

    \subsection{Dataset Availability}
    \label{appendix:dataset-availability}

    The \pedestrian dataset has been made publicly available to facilitate reproducibility and encourage further research in pedestrian safety~\cite{thomaPedestrianZenodo}.
    The complete dataset, including all video files and the balanced benchmark subset, is distributed as a 5.9~GB \texttt{zip} archive through Zenodo (DOI: \href{https://doi.org/10.5281/zenodo.10907945}{10.5281/zenodo.10907945}).
    Detailed instructions for accessing and using the dataset are provided in the accompanying \texttt{README} file included in the archive.

    \begin{figure}[htb]
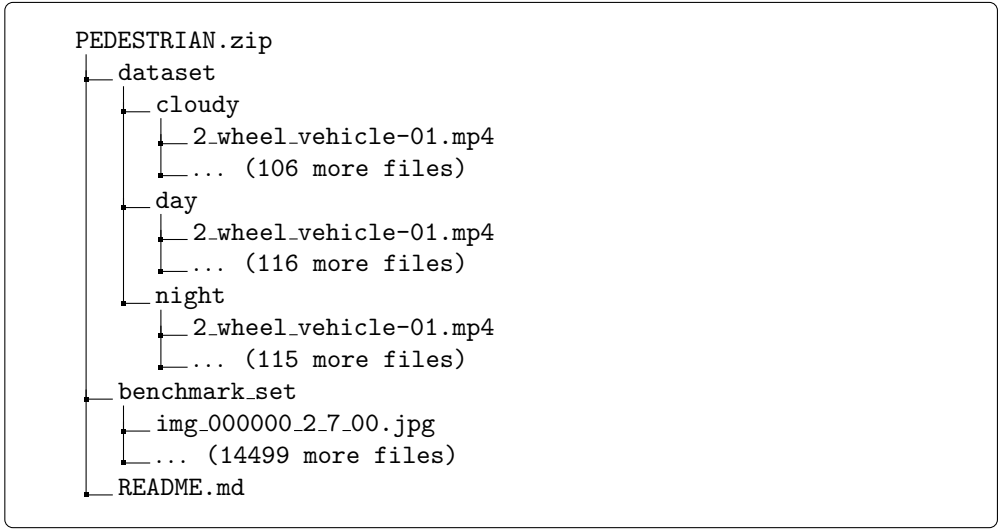

        \centering
        \begin{tcolorbox}[
                width=\textwidth,
                boxrule=0.4pt,
                colback=white,
                colframe=black
            ]
            \dirtree{%
                .1 PEDESTRIAN.zip.
                .2 dataset.
                .3 cloudy.
                .4 2\_wheel\_vehicle-01.mp4.
                .4 $\dots$ (106 more files).
                .3 day.
                .4 2\_wheel\_vehicle-01.mp4.
                .4 $\dots$ (116 more files).
                .3 night.
                .4 2\_wheel\_vehicle-01.mp4.
                .4 $\dots$ (115 more files).
                .2 benchmark\_set.
                .3 img\_000000\_2\_7\_00.jpg.
                .3 $\dots$ (14499 more files).
                .2 README.md.
            }
        \end{tcolorbox}
        \caption{The directory structure of the dataset.}
        \label{fig:dataset-structure}
    \end{figure}

    The dataset's directory structure is shown in Figure~\ref{fig:dataset-structure}.
    The \texttt{dataset/} subfolder contains the full \pedestrian dataset organized by lighting condition (\texttt{cloudy/}, \texttt{day/}, \texttt{night/}), with each video file stored in its respective condition folder.
    The \texttt{benchmark\_set/} subfolder contains a balanced subset of the dataset in terms of \taxLevel{Obstacle type} (see Section~\ref{sec:balanced-dataset-subset} for more information), as individual frames extracted from the videos.

    \newpage

    \subsection{Benchmark Experiments}
    \label{appendix:benchmark-experiments}
    All the relevant source code for working with the dataset is publicly available at \href{https://github.com/CYENS/PEDESTRIAN}{this GitHub repository}, which includes:

    \begin{itemize}
        \item Source code that can be used to replicate the benchmark experiments presented in Section~\ref{sec:benchmarks}.
        \item Implementations of PyTorch \texttt{Dataset} classes, to allow for easy loading and integration of the dataset into existing PyTorch training pipelines.
    \end{itemize}

    \section{Detailed Results}\label{appendix:detailed-results}

    Tables~\ref{tab:acc-obstacle-type-detailed-part1}~\&~\ref{tab:acc-obstacle-type-detailed-part2} show test accuracy results for all 16 DL models per \taxLevel{Obstacle type}.

    \begin{table*}[htb]
        \centering
        \caption{Test accuracy per obstacle type for the first 8 DL models (F: Frozen layers, NF: Not-frozen layers).} \label{tab:acc-obstacle-type-detailed-part1}
        \resizebox{1\textwidth}{!}{
            \begin{tabular}{lrrrrrrrrrrrrrrrr}
\toprule
& \multicolumn{2}{c}{\rotatebox{90}{Convnext Base}} & \multicolumn{2}{c}{\rotatebox{90}{Convnext Large}} & \multicolumn{2}{c}{\rotatebox{90}{Convnext Small}} & \multicolumn{2}{c}{\rotatebox{90}{Densenet121}} & \multicolumn{2}{c}{\rotatebox{90}{Densenet201}} & \multicolumn{2}{c}{\rotatebox{90}{Efficientnet B0}} & \multicolumn{2}{c}{\rotatebox{90}{Efficientnet V2 S}} & \multicolumn{2}{c}{\rotatebox{90}{Googlenet}} \\ \cmidrule(r){2-3} \cmidrule(r){4-5} \cmidrule(r){6-7} \cmidrule(r){8-9} \cmidrule(r){10-11} \cmidrule(r){12-13} \cmidrule(r){14-15} \cmidrule(r){16-17} 
\multirow[b]{-2}{*}{Obstacle Type} & F & NF & F & NF & F & NF & F & NF & F & NF & F & NF & F & NF & F & NF \\
\midrule
2-Wheel Vehicle & 1.000 & 1.000 & 1.000 & 1.000 & 1.000 & 1.000 & 1.000 & 1.000 & 1.000 & 1.000 & 1.000 & 1.000 & 1.000 & 1.000 & 1.000 & 1.000 \\
4-Wheel Vehicle & 1.000 & 1.000 & 1.000 & 1.000 & 1.000 & 1.000 & 1.000 & 0.998 & 1.000 & 1.000 & 1.000 & 1.000 & 0.988 & 1.000 & 1.000 & 1.000 \\
Advert Sign & 1.000 & 1.000 & 1.000 & 1.000 & 1.000 & 0.996 & 1.000 & 1.000 & 1.000 & 1.000 & 1.000 & 1.000 & 0.996 & 1.000 & 0.996 & 1.000 \\
Bench & 1.000 & 1.000 & 1.000 & 1.000 & 1.000 & 1.000 & 1.000 & 1.000 & 1.000 & 1.000 & 1.000 & 1.000 & 1.000 & 1.000 & 1.000 & 1.000 \\
Bin & 1.000 & 1.000 & 1.000 & 1.000 & 1.000 & 1.000 & 1.000 & 1.000 & 1.000 & 1.000 & 1.000 & 1.000 & 1.000 & 1.000 & 1.000 & 1.000 \\
Boulder & 1.000 & 1.000 & 1.000 & 1.000 & 1.000 & 1.000 & 1.000 & 1.000 & 1.000 & 1.000 & 0.996 & 1.000 & 1.000 & 1.000 & 1.000 & 1.000 \\
Bus Stop & 1.000 & 0.996 & 1.000 & 1.000 & 1.000 & 1.000 & 1.000 & 1.000 & 1.000 & 0.996 & 1.000 & 1.000 & 1.000 & 1.000 & 1.000 & 0.996 \\
Chair & 1.000 & 1.000 & 1.000 & 1.000 & 1.000 & 1.000 & 1.000 & 0.998 & 1.000 & 1.000 & 1.000 & 1.000 & 1.000 & 1.000 & 1.000 & 1.000 \\
Crack & 1.000 & 1.000 & 1.000 & 1.000 & 1.000 & 0.998 & 0.998 & 1.000 & 1.000 & 1.000 & 1.000 & 1.000 & 1.000 & 1.000 & 1.000 & 1.000 \\
Crowded Pavement & 1.000 & 1.000 & 1.000 & 1.000 & 1.000 & 1.000 & 1.000 & 1.000 & 1.000 & 1.000 & 1.000 & 1.000 & 1.000 & 1.000 & 1.000 & 1.000 \\
Fence & 1.000 & 1.000 & 1.000 & 1.000 & 1.000 & 1.000 & 1.000 & 1.000 & 1.000 & 1.000 & 1.000 & 1.000 & 1.000 & 1.000 & 1.000 & 1.000 \\
Flower & 1.000 & 0.996 & 1.000 & 1.000 & 1.000 & 1.000 & 1.000 & 1.000 & 1.000 & 1.000 & 1.000 & 1.000 & 0.996 & 1.000 & 0.976 & 1.000 \\
Hole/Pot-Hole & 1.000 & 0.996 & 1.000 & 0.995 & 1.000 & 0.996 & 1.000 & 1.000 & 1.000 & 1.000 & 1.000 & 1.000 & 0.996 & 1.000 & 0.996 & 0.996 \\
Information Tourist Sign & 1.000 & 1.000 & 1.000 & 1.000 & 1.000 & 1.000 & 1.000 & 1.000 & 1.000 & 1.000 & 1.000 & 1.000 & 1.000 & 1.000 & 1.000 & 1.000 \\
Light Fixture & 1.000 & 0.992 & 1.000 & 1.000 & 1.000 & 0.998 & 1.000 & 1.000 & 0.995 & 0.992 & 0.996 & 0.992 & 0.992 & 0.996 & 0.992 & 1.000 \\
Litter & 1.000 & 0.992 & 1.000 & 0.995 & 1.000 & 0.996 & 0.996 & 0.996 & 0.995 & 0.996 & 0.996 & 0.996 & 0.996 & 0.996 & 0.992 & 0.996 \\
Mail Box & 1.000 & 1.000 & 1.000 & 1.000 & 1.000 & 1.000 & 1.000 & 1.000 & 0.995 & 1.000 & 1.000 & 1.000 & 1.000 & 1.000 & 1.000 & 1.000 \\
Narrow Pavement & 1.000 & 1.000 & 1.000 & 1.000 & 1.000 & 1.000 & 1.000 & 1.000 & 1.000 & 1.000 & 1.000 & 1.000 & 1.000 & 1.000 & 1.000 & 1.000 \\
No Pavement & 1.000 & 1.000 & 1.000 & 1.000 & 1.000 & 0.998 & 0.996 & 1.000 & 0.995 & 1.000 & 0.992 & 1.000 & 1.000 & 0.996 & 0.984 & 1.000 \\
Parking Meter & 1.000 & 1.000 & 1.000 & 1.000 & 1.000 & 1.000 & 1.000 & 1.000 & 1.000 & 1.000 & 1.000 & 1.000 & 1.000 & 1.000 & 0.996 & 1.000 \\
Parking Prev. Barrier & 1.000 & 1.000 & 1.000 & 1.000 & 1.000 & 1.000 & 1.000 & 1.000 & 1.000 & 1.000 & 1.000 & 1.000 & 0.996 & 1.000 & 0.996 & 1.000 \\
Paver Broken & 1.000 & 1.000 & 1.000 & 1.000 & 1.000 & 0.998 & 1.000 & 1.000 & 1.000 & 1.000 & 0.996 & 1.000 & 0.992 & 1.000 & 1.000 & 1.000 \\
Plant Pot & 0.996 & 0.996 & 1.000 & 1.000 & 0.998 & 0.998 & 0.998 & 1.000 & 1.000 & 0.996 & 0.996 & 0.996 & 0.996 & 1.000 & 0.992 & 0.996 \\
Safety Sign & 1.000 & 0.996 & 1.000 & 1.000 & 1.000 & 0.996 & 0.998 & 0.993 & 1.000 & 1.000 & 1.000 & 0.996 & 0.996 & 1.000 & 0.992 & 0.988 \\
Scaffolding & 1.000 & 1.000 & 1.000 & 1.000 & 1.000 & 1.000 & 1.000 & 1.000 & 1.000 & 1.000 & 1.000 & 1.000 & 1.000 & 1.000 & 1.000 & 1.000 \\
Shrub & 1.000 & 0.996 & 1.000 & 1.000 & 1.000 & 1.000 & 1.000 & 1.000 & 1.000 & 1.000 & 1.000 & 1.000 & 0.988 & 1.000 & 0.988 & 1.000 \\
Table & 1.000 & 1.000 & 1.000 & 1.000 & 1.000 & 1.000 & 1.000 & 1.000 & 1.000 & 1.000 & 1.000 & 1.000 & 1.000 & 1.000 & 0.992 & 0.996 \\
Traffic Cone & 1.000 & 1.000 & 1.000 & 1.000 & 1.000 & 1.000 & 1.000 & 1.000 & 1.000 & 1.000 & 1.000 & 1.000 & 1.000 & 1.000 & 1.000 & 1.000 \\
Tree & 1.000 & 1.000 & 1.000 & 1.000 & 1.000 & 1.000 & 1.000 & 1.000 & 1.000 & 1.000 & 1.000 & 1.000 & 0.996 & 1.000 & 0.996 & 1.000 \\
\bottomrule
\end{tabular}

        }
    \end{table*}

    \begin{table*}[htb]
        \centering
        \caption{Test accuracy per obstacle type for the last 8 DL models (F: Frozen layers, NF: Not-frozen layers).} \label{tab:acc-obstacle-type-detailed-part2}
        \resizebox{1\textwidth}{!}{
            \begin{tabular}{lrrrrrrrrrrrrrrrr}
\toprule
& \multicolumn{2}{c}{\rotatebox{90}{Mobilenet V2}} & \multicolumn{2}{c}{\rotatebox{90}{Mobilenet V3 Large}} & \multicolumn{2}{c}{\rotatebox{90}{Mobilenet V3 Small}} & \multicolumn{2}{c}{\rotatebox{90}{Resnet101}} & \multicolumn{2}{c}{\rotatebox{90}{Resnet152}} & \multicolumn{2}{c}{\rotatebox{90}{Resnet18}} & \multicolumn{2}{c}{\rotatebox{90}{Resnet34}} & \multicolumn{2}{c}{\rotatebox{90}{Resnet50}} \\ \cmidrule(r){2-3} \cmidrule(r){4-5} \cmidrule(r){6-7} \cmidrule(r){8-9} \cmidrule(r){10-11} \cmidrule(r){12-13} \cmidrule(r){14-15} \cmidrule(r){16-17} 
\multirow[b]{-2}{*}{Obstacle Type} & F & NF & F & NF & F & NF & F & NF & F & NF & F & NF & F & NF & F & NF \\
\midrule
2-Wheel Vehicle & 1.000 & 1.000 & 1.000 & 1.000 & 1.000 & 1.000 & 1.000 & 1.000 & 1.000 & 1.000 & 1.000 & 0.995 & 1.000 & 0.995 & 1.000 & 1.000 \\
4-Wheel Vehicle & 1.000 & 0.995 & 1.000 & 1.000 & 0.995 & 0.995 & 1.000 & 0.990 & 1.000 & 0.995 & 0.990 & 0.995 & 1.000 & 1.000 & 1.000 & 0.985 \\
Advert Sign & 1.000 & 1.000 & 1.000 & 1.000 & 1.000 & 0.995 & 1.000 & 0.995 & 1.000 & 0.995 & 0.995 & 0.995 & 1.000 & 0.995 & 1.000 & 1.000 \\
Bench & 1.000 & 0.990 & 1.000 & 1.000 & 1.000 & 1.000 & 1.000 & 1.000 & 0.995 & 1.000 & 1.000 & 1.000 & 1.000 & 1.000 & 1.000 & 0.995 \\
Bin & 1.000 & 1.000 & 1.000 & 1.000 & 0.995 & 1.000 & 1.000 & 1.000 & 1.000 & 1.000 & 1.000 & 1.000 & 1.000 & 1.000 & 1.000 & 1.000 \\
Boulder & 1.000 & 0.995 & 1.000 & 1.000 & 1.000 & 1.000 & 1.000 & 1.000 & 1.000 & 1.000 & 0.995 & 1.000 & 1.000 & 0.995 & 1.000 & 1.000 \\
Bus Stop & 1.000 & 0.995 & 0.995 & 1.000 & 1.000 & 1.000 & 0.995 & 1.000 & 1.000 & 0.995 & 1.000 & 1.000 & 1.000 & 0.995 & 1.000 & 1.000 \\
Chair & 1.000 & 1.000 & 1.000 & 1.000 & 1.000 & 0.995 & 0.995 & 1.000 & 0.995 & 0.990 & 1.000 & 1.000 & 0.995 & 1.000 & 1.000 & 0.995 \\
Crack & 1.000 & 0.995 & 1.000 & 0.995 & 1.000 & 1.000 & 1.000 & 0.995 & 0.995 & 1.000 & 0.995 & 1.000 & 1.000 & 1.000 & 1.000 & 1.000 \\
Crowded Pavement & 1.000 & 1.000 & 1.000 & 1.000 & 1.000 & 1.000 & 1.000 & 1.000 & 1.000 & 1.000 & 1.000 & 1.000 & 1.000 & 1.000 & 1.000 & 1.000 \\
Fence & 1.000 & 1.000 & 0.985 & 1.000 & 1.000 & 1.000 & 1.000 & 1.000 & 1.000 & 1.000 & 1.000 & 1.000 & 1.000 & 1.000 & 1.000 & 1.000 \\
Flower & 1.000 & 1.000 & 1.000 & 1.000 & 1.000 & 0.995 & 1.000 & 1.000 & 1.000 & 1.000 & 0.995 & 1.000 & 1.000 & 1.000 & 1.000 & 1.000 \\
Hole/Pot-Hole & 1.000 & 0.995 & 1.000 & 0.995 & 0.995 & 0.995 & 0.995 & 1.000 & 1.000 & 1.000 & 0.995 & 1.000 & 1.000 & 1.000 & 1.000 & 1.000 \\
Information Tourist Sign & 0.995 & 1.000 & 1.000 & 1.000 & 1.000 & 1.000 & 1.000 & 1.000 & 1.000 & 0.995 & 1.000 & 1.000 & 1.000 & 1.000 & 1.000 & 1.000 \\
Light Fixture & 0.995 & 0.995 & 1.000 & 1.000 & 0.995 & 1.000 & 0.990 & 1.000 & 1.000 & 1.000 & 0.995 & 0.995 & 0.995 & 0.990 & 1.000 & 1.000 \\
Litter & 0.995 & 0.995 & 0.995 & 0.995 & 0.995 & 0.995 & 0.995 & 0.995 & 1.000 & 0.995 & 0.995 & 0.995 & 0.995 & 0.995 & 0.995 & 0.995 \\
Mail Box & 1.000 & 1.000 & 1.000 & 1.000 & 1.000 & 0.995 & 1.000 & 1.000 & 1.000 & 0.995 & 0.995 & 1.000 & 1.000 & 1.000 & 0.995 & 1.000 \\
Narrow Pavement & 1.000 & 1.000 & 1.000 & 1.000 & 1.000 & 1.000 & 1.000 & 1.000 & 0.995 & 1.000 & 0.995 & 1.000 & 0.990 & 1.000 & 1.000 & 1.000 \\
No Pavement & 1.000 & 1.000 & 0.985 & 1.000 & 1.000 & 1.000 & 1.000 & 1.000 & 1.000 & 1.000 & 0.990 & 1.000 & 0.995 & 1.000 & 0.995 & 1.000 \\
Parking Meter & 1.000 & 0.995 & 0.985 & 1.000 & 1.000 & 1.000 & 0.995 & 1.000 & 0.995 & 1.000 & 1.000 & 1.000 & 1.000 & 1.000 & 0.990 & 1.000 \\
Parking Prev. Barrier & 1.000 & 1.000 & 1.000 & 1.000 & 0.995 & 1.000 & 0.995 & 1.000 & 1.000 & 1.000 & 1.000 & 1.000 & 1.000 & 1.000 & 0.995 & 1.000 \\
Paver Broken & 0.995 & 1.000 & 0.995 & 1.000 & 0.990 & 1.000 & 1.000 & 1.000 & 1.000 & 1.000 & 0.995 & 1.000 & 0.990 & 1.000 & 1.000 & 1.000 \\
Plant Pot & 0.995 & 1.000 & 1.000 & 1.000 & 1.000 & 1.000 & 0.990 & 1.000 & 1.000 & 1.000 & 0.995 & 0.990 & 0.985 & 1.000 & 0.995 & 0.995 \\
Safety Sign & 0.985 & 0.995 & 1.000 & 0.995 & 0.990 & 0.990 & 0.995 & 0.980 & 0.995 & 0.990 & 0.985 & 0.990 & 0.985 & 0.990 & 0.995 & 0.985 \\
Scaffolding & 1.000 & 1.000 & 1.000 & 0.995 & 1.000 & 1.000 & 1.000 & 1.000 & 1.000 & 0.995 & 1.000 & 0.995 & 1.000 & 1.000 & 1.000 & 1.000 \\
Shrub & 0.995 & 0.995 & 0.995 & 1.000 & 1.000 & 1.000 & 0.990 & 1.000 & 0.995 & 1.000 & 0.995 & 1.000 & 0.995 & 1.000 & 1.000 & 1.000 \\
Table & 1.000 & 1.000 & 1.000 & 1.000 & 1.000 & 1.000 & 0.995 & 1.000 & 1.000 & 1.000 & 1.000 & 1.000 & 1.000 & 1.000 & 1.000 & 1.000 \\
Traffic Cone & 1.000 & 1.000 & 1.000 & 1.000 & 1.000 & 1.000 & 1.000 & 1.000 & 1.000 & 1.000 & 1.000 & 1.000 & 1.000 & 1.000 & 1.000 & 1.000 \\
Tree & 1.000 & 0.995 & 1.000 & 1.000 & 1.000 & 1.000 & 1.000 & 0.995 & 1.000 & 1.000 & 1.000 & 1.000 & 0.995 & 0.995 & 1.000 & 1.000 \\
\bottomrule
\end{tabular}

        }
    \end{table*}

\end{appendices}

\pagebreak

\bibliography{bibliography}

\end{document}